\documentclass{article}

\PassOptionsToPackage{numbers}{natbib}
\usepackage[final]{neurips_2023}

\usepackage[utf8]{inputenc} 
\usepackage[T1]{fontenc}    
\usepackage{hyperref}       
\usepackage{url}            
\usepackage{booktabs}       
\usepackage{amsfonts}       
\usepackage{nicefrac}       
\usepackage{microtype}      
\usepackage{xcolor}         
\usepackage{lmodern}
\usepackage[pdftex]{graphicx}
\usepackage{amsmath}
\usepackage{cleveref}

\title{Interactive Visual Feature Search}

\author{%
  Devon Ulrich\thanks{Work done as a student at Princeton University} \\
  Jane Street\\
  \texttt{dulrich@alumni.princeton.edu} \\
  \And
  Ruth Fong\\
  Princeton University\\
  \texttt{ruthfong@cs.princeton.edu} \\
}

\begin{document}

\maketitle

\begin{abstract}
    Many visualization techniques have been created to explain the behavior of computer vision models, but they largely consist of static diagrams that convey limited information.
    Interactive visualizations allow users to more easily interpret a model's behavior, but most are not easily reusable for new models.
    We introduce Visual Feature Search, a novel interactive visualization that is adaptable to any CNN and can easily be incorporated into a researcher's workflow. 
    Our tool allows a user to highlight an image region and search for images from a given dataset with the most similar model features.
    We demonstrate how our tool elucidates different aspects of model behavior by performing experiments on a range of applications, such as in medical imaging and wildlife classification.
    \setcounter{footnote}{0}
    Our tool is open source and can be used by others to interpret their own models.\footnote{Our code is available at \href{https://github.com/lookingglasslab/VisualFeatureSearch}{\texttt{https://github.com/lookingglasslab/VisualFeatureSearch}}}
\end{abstract}

\section{Introduction}

Computer vision models such as convolutional neural networks (CNNs) and visual transformers are notoriously hard to interpret due to their size and complexity. Various techniques have been proposed to help visualize and ``explain'' these models with static figures; 
for instance, attribution heatmaps~\cite{bach2015pixel,fong17interpretable,simonyan14deep,smilkov2017smoothgrad,zhou2016learning} like Grad-CAM \cite{selvaraju17gradcam} visualize which input image regions are important for a model's output decision, and feature visualization techniques help explain internal aspects of models (e.g. what visual stimuli most activates a given neuron) \cite{netdissect2017,mahendran16visualizing,olah2017feature,simonyan14deep,zeiler2014visualizing}.

However, researchers have recently focused on creating \emph{interactive} visualizations of CNNs, which can present more data in an easy-to-use way. Several works \cite{Bau:Ganpaint:2019, carter2019activation, harley2015isvc, Lee_2022_CVPR, norton2017adversarial, olah2018building, microscope} provide graphical interfaces that allow the user to interact with CNNs and produce rich visualizations. While these tools are effective at explaining CNN behavior, they are generally only designed for a handful of pre-selected models. A key criteria for the adoption of interpretability techniques is how easy they are to incorporate into a researcher's workflow; significant effort is required to utilize these interactive tools in new experiments, so they are unfortunately not widely used in practice.

Some works, such as Teachable Machines \cite{webster2017now}, What If \cite{wexler2019if}, TensorBoard \cite{abadi2016tensorflow}, and Interactive Similarity Overlays (ISO) \cite{fong_interactive_2021} are more lightweight and easy to integrate with new models, but only the latter two can visualize internal feature data. TensorBoard only supports basic feature visualizations (i.e.~plotting distributions of activations) while ISO enables users to qualitatively compare spatial CNN features, but only for a handful of images at a time.

\begin{figure}[t]
  \centering
  \includegraphics[width=0.95\linewidth]{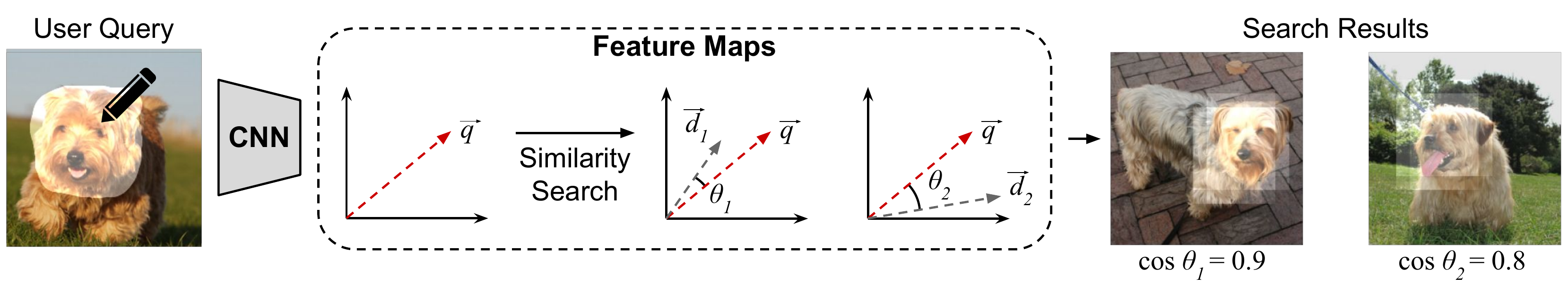}
  \caption{\emph{Overview}. The user highlights a region in the query image via our interactive tool (left); then, our tool searches through an image dataset and returns images with the most similar CNN features to that of the selected query region (right). The cosine similarity scores between returned image regions and the selected query region are shown. See Section~\ref{sec:approach} for more details.}
  \label{fig:overview}
\end{figure}

In this paper, we introduce Visual Feature Search (VFS), a novel interactive visualization that empowers machine learning researchers to easily explore the visual features of almost any computer vision model. Our tool is designed to be lightweight and flexible so that users can quickly set up VFS to analyze the intermediate features of arbitrary CNNs and visual transformers; the only requirement is that the model has intermediate spatial features (i.e.~a 3D tensor of shape $H \times W \times C$). Our visualization allows a user to highlight a free-form region in an image, and it searches for other images in a dataset that contain similar feature representations to the highlighted region and displays the most similar results (\cref{fig:overview}). 
In essesence, this provides a visual explanation to answer the question, ``what does the model consider to be most similar to this image region?'' To fully showcase the interactivity of VFS, we provide several demo videos and interactive Jupyter notebooks in addition to this paper (see supp.~mat.).

Our work is similar to CNN-based approaches that tackle content-based instance retrieval (CBIR), which aims to find visually similar images to a query image~\cite{cbir3,cbir2,zheng2017sift}. However, our works differs from CBIR methods in two ways: First, our goal is to understand a CNN (i.e. interpretability) by investigating what visual patterns are similar in \emph{feature space}, whereas instance retrieval focuses on retrieving \emph{visually similar} images (and often uses CNN features to do so). Second, we explicitly focus on leveraging \emph{interactivity} to allow users to quickly and iteratively gain insights on their selected models (e.g.~by experimenting with multiple ideas in quick succession). 

In the remainder of this paper, we summarize our implementation of VFS and include several experiments to highlight how it can be used to better understand computer vision models. 
Our source code and interactive Jupyter notebooks are available on GitHub;\footnote{\href{https://github.com/lookingglasslab/VisualFeatureSearch}{\texttt{https://github.com/lookingglasslab/VisualFeatureSearch}}} our goal is to enable other researchers and practitioners to use VFS as a new method for interpreting their models.

\section{Approach}
\label{sec:approach}
To use VFS, the user first selects a model and a layer within it to study, as well as a dataset of images to search across. We provide the user with an interactive widget for selecting a query image and highlighting a free-form region in it to use as the search query.
To perform a feature search across the dataset, our tool computes the feature maps of images immediately after the user's selected layer and compares the highlighted regions within these maps. 

Formally, let $f_l(\mathbf{q}) \in \mathbb{R}^{H \times W \times C}$ be the $l$-th layer's feature map for the query image $\mathbf{q}$. If we down-sample the user's selected region of $\mathbf{q}$ into a mask $\mathbf{m}_l \in [0,1]^{H \times W}$, we can apply the mask to obtain a 3D tensor $\mathbf{z} \in \mathbb{R}^{H \times W \times C}$ s.t.~$ \mathbf{z}_{(i,j,k)} := f_l(\mathbf{q})_{(i,j,k)} \cdot \mathbf{m}_{l(i,j)} $.
To convert this into a query vector $\vec{q}$ for similarity search, we crop $\mathbf{z}$ to remove any zero padding and flatten the resulting data into a vector.
We use a similar process to convert the feature maps of all images in the search dataset into vectors. We apply the mask $\mathbf{m}_l$ as a sliding window over each image in the search dataset to create region vectors $\vec{d}_i$ with the same dimensions as $\vec{q}$. This allows us to compare the query vector to each $\vec{d}_i$ via cosine similarities; we sort all search regions by their similarity scores and display the most similar image regions to the user (\cref{fig:overview}), thereby visualizing images with the most similar intermediate features to the user's selected region. 

In order to use this algorithm for large-scale, real time searches, we precompute the features $f_l(\mathbf{d}_i)$ for all dataset images $\mathbf{d}_i$, and we store the resulting data in a compressed cache file via the Zarr Python library \cite{alistair_miles_2020_3773450}. The cache file allows VFS experiments to be easily shared and reproduced between multiple users, such as by downloading the file and running VFS on Google Colab environments. Furthermore, if the user wishes to search across a large dataset with features that cannot be stored in-memory (e.g. $>50,000$ images with ResNet50 conv5 features), then VFS can load features from the cache file to efficiently compute search results. Additionally, VFS is implemented with several GPU optimizations in PyTorch \cite{NEURIPS2019_9015} in order to compute results in real time (see supp.~mat.~for more details).

\section{Experiments and Demonstrations}

\paragraph{Domain Generalization.}
One application of VFS is to understand how robust a model is when presented with novel images. To demonstrate this, we visualize ResNet50~\cite{he2016deep} conv5 features of in- and out-of-domain (o.o.d.) images in two sets of experiments; our goal is to investigate whether a model's internal feature representations of in-domain images is similar to those of o.o.d.~images. 

\begin{figure}
    \centering
    \includegraphics[width=0.95\linewidth]{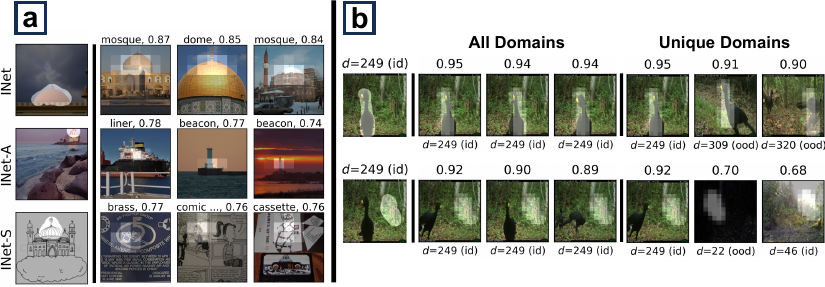}
    \caption{\emph{Domain Generalization}. Each row contains one query image on the left, followed by multiple VFS results and similarity scores. 
    \textbf{a}: Three queries of mosque images from ImageNet, ImageNet-A \cite{hendrycks2021nae}, and ImageNet-Sketch \cite{wang2019learning} on an ImageNet-trained model. 
    \textbf{b}: queries and results from the iWildCam dataset \cite{beery2020iwildcam} on a pretrained classifier. Results include the top-3 overall results and the top-3 results for unique domains. 
    The ImageNet model struggles to featurize o.o.d.~mosque images, while the iWildCam model produces very generalizable animal features.}
    \label{fig:out-of-domain}
\end{figure}

The first experiments search for the most similar images in the ImageNet validation set \cite{russakovsky2015imagenet} when similar query regions are selected from images in the ImageNet test set (in-domain), ImageNet-A dataset \cite{hendrycks2021nae}, and ImageNet-Sketch dataset \cite{wang2019learning} (both o.o.d.). One example is shown in \Cref{fig:out-of-domain}a, where images of mosques are selected as query images from all three datasets. The nearest neighbor results for all three queries show that the model can accurately extract semantic data from the in-domain query image, but it fails to encode the other two queries as mosques due to their out-of-distribution scale and texture; additionally, the cosine similarities are much higher for the in-domain query as opposed to the two o.o.d.~queries. Additional queries corroborate these trends (see supp.~mat.).

The second set of experiments use the iWildCam dataset \cite{beery2020iwildcam}, which consists of images of wildlife from various trap camera locations. Some locations are included in the training subset while some are withheld and are thus out-of-domain. We investigate the conv5 features from a ResNet50 model that was trained to detect the presence of animals and classify their species \cite{koh2021wilds}; representative VFS queries and results are shown in \Cref{fig:out-of-domain}b. 
These results support the finding in \cite{koh2021wilds} that the model is able to generalize fairly well, as the feature representations for animals have high similarity scores (e.g.~0.9) across different domains;
in contrast, when a background patch of an image is queried, only images from the same domain have similarity scores above 0.7. 

\paragraph{Chest X-ray Classifiers.}

Another use of VFS is to understand why a model made a particular decision by finding image regions that correlate with certain classification labels. To demonstrate this, we turn to the domain of chest X-ray classification: we study the last feature layer of a pretrained DenseNet-121 \cite{Cohen2022xrv,huang2018densely} that classifies pathologies in the ChestXray-14 dataset \cite{Wang_2017_ChestX}. 

\begin{figure}
    \centering
    \includegraphics[width=0.95\linewidth]{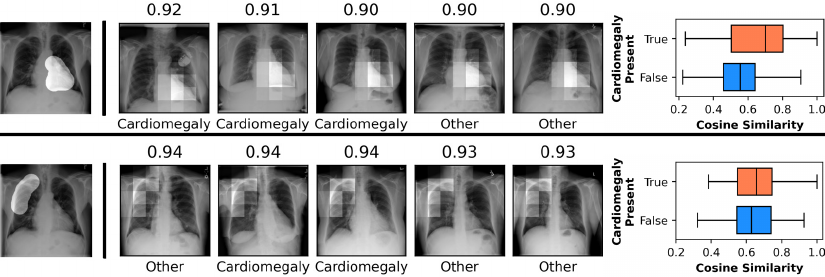}
    \caption{\emph{Interpreting Chest X-ray Classification.} Queries and results of a patient X-ray with cardiomegaly from the ChestXray-14 dataset \cite{Wang_2017_ChestX}. The top-5 results are shown for two queries (left) as well as box plots displaying the distribution of cosine similarities between the query regions and all dataset images (right). 
    While the qualitative results between the two queries only differ slightly, the box plots show that the similarity scores of images containing cardiomegaly tend to be higher than those without specifically when the query region contains the heart.}
    \label{fig:xray}
\end{figure}

We specifically investigate an X-ray image of a patient with cardiomegaly (i.e. condition of having an enlarged heart). The model is able to correctly classify this patient as having cardiomegaly, and we use the same X-ray image as a query in VFS to provide interpretable visualizations of the model features. Our results are shown in \Cref{fig:xray}: we find that highlighting the heart of the patient returns nearest neighbor regions of other hearts from patients with cardiomegaly; in contrast, searching for an unrelated region of the same patient's X-ray (e.g. lung) yields nearest neighbors with mixed diagnoses. Empirically, when the heart is highlighted, images of patients with cardiomegaly tend to have higher similarity scores than images of patients without the condition; in contrast, when the lung is selected as the search query, the distribution of similarity scores is virtually identical for X-rays with and without cardiomegaly.
This suggests that the classifier's prediction of cardiomegaly is correlated specifically with the heart region of the X-ray.

\paragraph{Editing Classifiers.}
Recently, a method to correct for systematic CNN mistakes was introduced \cite{santurkar2021editing}. For instance, \cite{santurkar2021editing} found that a VGG16 ImageNet classifier consistently misclassifies vehicles that are on a snowy surface and typically predicts these images as snowmobiles or snowplows.
Their CNN-editing method mitigated this mistake by updating the model's weights such that the model treats snowy terrain as if it were asphalt. We use VFS to explore the original and edited VGG models in the ``vehicles on snow'' example and visualize how the edit affected the model's features by analyzing nearest neighbor search results from the 
ImageNet validation set.

\begin{figure}[h]
    \centering
    \includegraphics[width=0.95\linewidth]{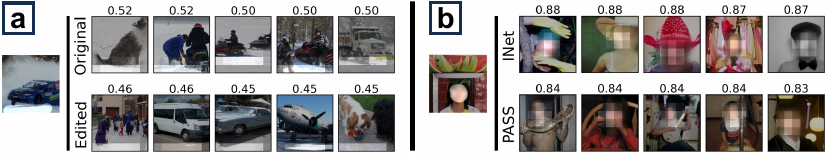}
    \caption[Caption]{\emph{Editing a Classifier and Training on PASS}. \textbf{a}: Editing a classifier to improve accuracy for vehicles on snow.  The top rows show that the original model encoded the ground surface similarly to ice and snow, while the bottom row shows that the edited model more generally relates the ground to asphalt. \textbf{b}: Facial feature data for an ImageNet- vs.~PASS-trained model.\footnotemark  Despite not being trained on images containing humans, the PASS model encodes features that can match the query face to other faces in the dataset.}
    \label{fig:editing-pass}
\end{figure}

Our visualizations suggest that the snow-to-asphalt edit worked and has a noticeable effect on intermediate features. We highlight an example of the edit in \Cref{fig:editing-pass}a. Our search results show that the original model does not have a clear understanding that the ground should be treated like asphalt road; the nearest neighbors include other surfaces such as snow, and ice. In contrast, the search results from the edited model include asphalt surfaces and several cars (as opposed to snowmobiles and snowplows in original search results), which indicates that the model edit successfully changed the feature representation of snowy roads to achieve the desired result.

\paragraph{ImageNet vs. PASS.}

We next study how the choice of training dataset affects a model's feature representation. To mitigate privacy concerns of training on images with humans, \cite{asano21pass} introduced the PASS, a dataset of unlabeled images that do not contain human faces or body parts. PASS is meant serve as an ImageNet replacement for self-supervised learning and has been shown to perform as well as ImageNet-trained models on human-centric tasks (e.g.~pose estimation). 
We compare the feature representations from two ResNet50 models, one trained on ImageNet and the other on PASS, that were trained via MoCo-v2 \cite{chen2020mocov2} self-supervision.
Our results when using VFS on both models are included in \Cref{fig:editing-pass}b. The most notable observation is that the PASS-trained model is able to accurately match face queries to other faces in the dataset, despite never being trained with images of humans.
However, the similarity scores for the ImageNet model results are generally greater than those for the PASS model.


\newpage

\section{Conclusion}
In summary, we propose a new interactive tool for understanding the intermediate activations of CNNs.
Many existing interactive visualizations can not be easily applied to new models and/or datasets; thus, they are often not utilized by others as regular research tools.
We demonstrate through our experiments that our tool is much more flexible in comparison: it can be used to quickly visualize new models and datasets, and we hope that this flexibility allows other researchers to use it to better understand their own models.
Lastly, we emphasize that our tool is qualitative and should be paired with quantitative analysis to fully corroborate findings.

\footnotetext{Faces in this figure are blurred to preserve the privacy of the photographed individuals.}

\begin{ack}
We are grateful for support from Open Philanthropy (RF), the Princeton Engineering Project X Fund (RF), and the Princeton SEAS IW Funding (DU).
We thank David Bau, Sunnie S. Y. Kim, and Indu Panigrahi for helpful discussions and/or feedback on our tool.
We also thank the authors of~\cite{Bau:Ganpaint:2019}, whose widget we based our highlighting widget on.
We are grateful to the authors of~\cite{asano21pass, opencv_library, chen2020mocov2, Cohen2022xrv, koh2021wilds, alistair_miles_2020_3773450, NEURIPS2019_9015, santurkar2021editing} for open-sourcing their code and/or trained models. 
\end{ack}

{
\small
\bibliography{references, fong19-refs, fong20-thesis-refs}
}


\newpage
\appendix
\section{Supplementary Materials}

\paragraph{Code and Demos}
Our source code and sample Jupyter Notebooks that showcase VFS are available at {\href{https://github.com/lookingglasslab/VisualFeatureSearch}{\texttt{https://github.com/lookingglasslab/VisualFeatureSearch}}}. The notebooks are designed to run in Google Colab, and the repository includes additional instructions for running VFS locally with Jupyter. 

\paragraph{Additional Figures}

In this section, we provide additional figures for the domain generalization, editing classifiers, and ImageNet vs.~PASS experiments.

\begin{figure}[b]
    \centering
    \includegraphics[width=0.95\linewidth]{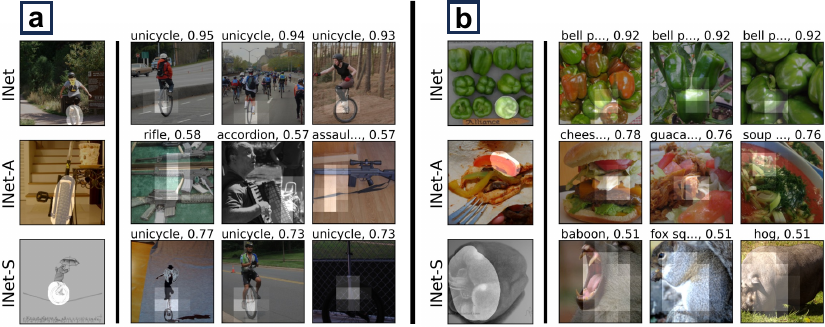}
    \caption{\emph{Additional ImageNet Generalization.} Three queries of images from ImageNet, ImageNet-A \cite{hendrycks2021nae}, and ImageNet-Sketch \cite{wang2019learning} on an ImageNet model. \textbf{a}: All queries are of unicycle wheels. The ImageNet-Sketch unicycle is matched with other unicycle wheels in the dataset, but the similarity scores are lower than those from the in-domain query. \textbf{b}: All queries are of bell peppers, but only the in-domain search contains bell peppers in the results. Such visuals may provide insights into why a particular model misclassified a challenging example.}
    \label{fig:supp-ood}
\end{figure}

Figure \ref{fig:supp-ood} includes additional queries and results for the ImageNet (in-domain), ImageNet-A, and ImageNet-Sketch (o.o.d.) experiments. Fig.~\ref{fig:supp-ood}a, all queries are pictures of unicycles; when the unicycle wheels are highlighted in each query, both the ImageNet and ImageNet-Sketch queries are successfully matched to other unicycle wheels, while the ImageNet-A query is matched with various unrelated nearest neighbors. Similarly, in Fig. \ref{fig:supp-ood}b, all three queries are images of bell peppers; however, only the in-domain query yields nearest neighbors that are also bell peppers. For both the unicycle and the bell pepper queries, the resulting similarity scores are highest for the in-domain queries (i.e.~$> 0.9$) when compared to the scores for the o.o.d. queries (i.e.~$<0.8$).

\begin{figure}
    \centering
    \includegraphics[width=0.95\linewidth]{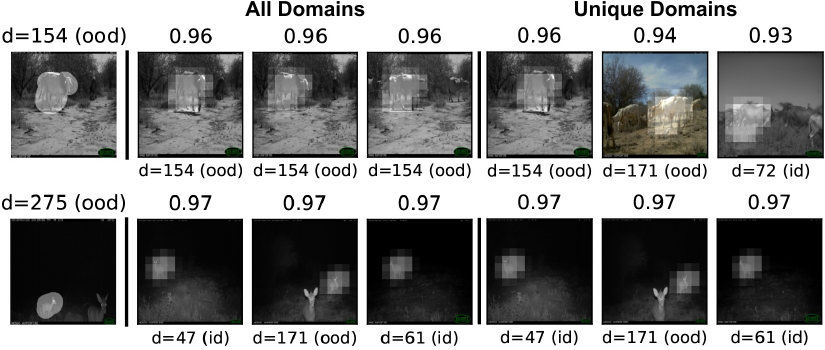}
    \caption{\emph{Additional iWildCam Generalization.} Two sets of queries and results for a cow (top) and a deer (bottom). While the deer features appear to be highly generalizable with all top-3 results originating from different camera locations, the cow has a comparatively worse feature representation since its nearest neighbors from other domains have slightly lower similarity scores (0.94, 0.93) than inner-domain results (0.96); additionally, its nearest neighbors from other domains are images of horses, not cattle.}
    \label{fig:sup-wilds}
\end{figure}

Similarly, Figure \ref{fig:sup-wilds} includes additional domain generalization visuals for the iWildCam dataset. Two sets of queries and search results are shown: one is of a cow during the day, while the other is of a deer at night. The results for the deer query are similar to those in Figure \ref{fig:out-of-domain}, as the features appear to be highly generalizable have have nearest neighbors of other deer across a variety of domains. However, the encoded features for the cow are less generalizable and simultaneously less accurate, as the nearest neighbors in other domains have lower similarity scores (0.94, 0.93) than those from the same domain (0.96), and the nearest neighbor images from other domains actually contain horses, not cows. This particular query image is misclassified by the model as containing a horse, so the search results help visualize the features associated with this misclassification. 

\begin{figure}
    \centering
    \includegraphics[width=0.95\linewidth]{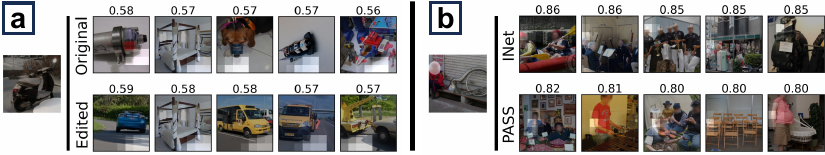}
    \caption{\emph{Additional Results for Edited Classifier and PASS Training}. \textbf{a}: An additional visualization of the ``vehicles on snow'' classifier edit. The original model's VFS results consist entirely of unrelated objects such as ice, floors, and a tabletop, whereas the edited model's  results contain several cars on asphalt. \textbf{b}: An additional search query and results for ImageNet- vs.~PASS-trained models. While the localization of nearest neighbors is relatively poor for PASS (i.e. the highlighted regions in some search results do not contain faces), the PASS model is able to successfully able to match the query to other images that contain faces.}
    \label{fig:supp-editing-pass}
\end{figure}

Figure \ref{fig:supp-editing-pass} contains an additional example of the Editing Classifiers visualization, as well as an additional visual for the ImageNet vs.~PASS experiment. Fig.~\ref{fig:supp-editing-pass}a includes a query of a scooter on snow-covered ground; when the ground is highlighted, the original model's VFS results contain no other images of scooters or cars. However, when the edited model is used, four of the top-5 results contain cars, and the the instance of a bobsled on ice from the original results is no longer in the top-5 results. Thus, this example provides further evidence that the model edit was successful. 

Figure \ref{fig:supp-editing-pass}b shows an additional example of a query containing a face with two sets of results for the ImageNet and PASS models, respectively. Although several of the highlighted regions in the PASS results contain no faces, two such results contain faces elsewhere in the image. Thus, the PASS-trained model is again able to successfully encode human faces and retrieve other images containing faces via VFS.

\paragraph{Implementation Details}
In Section \ref{sec:approach}, we described how to perform the region-based similarity search for VFS. However, in order to perform the search efficiently, we use a modified version of this algorithm that substitutes the sliding window approach with a convolution operation. This allows us to compute the searches on a GPU via PyTorch, which allows for extensive parallelization and enables each search to be much faster than if it were run on a CPU. 

To implement the convolution-based searches, we first take the 3D tensor $\mathbf{z} \in \mathbb{R}^{H\times W \times C}$ from Section \ref{sec:approach} and apply the mask $\mathbf{m}_l$ on it once more. We then crop the tensor, which we define as removing all rows/columns on the exterior of the feature map which only contain zero-valued elements. Let the resulting tensor be $\mathbf{z'}$, where:
\begin{align}
    \mathbf{z}'_{(i',j',k')} = \text{Crop}( \mathbf{z}_{(i,j,k)} \cdot \mathbf{m}_{l(i,j)})
\end{align}
$\mathbf{z}'$ has dimensions $H' \times W' \times C$, where $H' \leq H$ and $W' \leq W$. We use $\mathbf{z}'$ as a 2D convolutional filter and apply it to a feature map $f_l(\mathbf{s})$ from the search database.
\begin{align}
    \mathbf{c} := f_l(\mathbf{s}) * \mathbf{z}'
\end{align}
Each element $\mathbf{c}_{(a,b)}$ is equivalent to the inner product $\vec{q} \cdot \vec{r}_i$, for a unique region vector $\vec{r}_i$ within the search image's feature map. We can perform a similar convolution to obtain the magnitudes of each $\vec{r_i}$, so we can thus compute the cosine similarities for all searchable regions within the image $\mathbf{s}$ without iteratively computing results with a sliding window.

\end{document}